\documentclass[11pt,letterpaper]{article}
\usepackage{emnlp2016}
\usepackage{times}
\usepackage{latexsym}
\usepackage{graphicx}
\usepackage{epstopdf}
\usepackage{amssymb}
\usepackage{booktabs}
\usepackage{multirow}
\usepackage{amsmath}

\emnlpfinalcopy

\title{An Investigation of Recurrent  Neural  Architectures \\ for Drug Name Recognition}

 \author{Raghavendra  Chalapathy \\ University of Sydney\\
 	J12/1  Cleveland St \\ Darlington NSW 2008 \\ rcha9612@uni.sydney.edu.au
         \And  
         Ehsan Zare Borzeshi  \\ Capital Markets CRC\\
         3/55 Harrington St \\Sydney NSW 2000 \\ ezborzeshi@cmcrc.com
         \And
         Massimo Piccardi \\University of Technology Sydney \\ PO Box 123 \\  Broadway NSW 2007 \\
         Massimo.Piccardi@uts.edu.au
         }

\begin{document}

\maketitle

\begin{abstract}
Drug name recognition (DNR) is an essential step in the Pharmacovigilance (PV) pipeline. DNR aims to find drug name mentions in unstructured biomedical texts and classify them into predefined categories. State-of-the-art DNR approaches heavily rely on hand-crafted features and domain-specific resources which are difficult to collect and tune. For this reason, this paper investigates the effectiveness of contemporary recurrent neural architectures - the Elman and Jordan networks and the bidirectional LSTM with CRF decoding - at performing DNR straight from the text. The experimental results achieved on the authoritative SemEval-2013 Task 9.1 benchmarks show that the bidirectional LSTM-CRF ranks closely to highly-dedicated, hand-crafted systems.
\end{abstract}

\section{Introduction}
Pharmacovigilance (PV) is defined by the World Health Organization as the science and activities concerned with the detection, assessment, understanding and prevention of adverse effects of drugs or any other drug-related problems. Drug name recognition (DNR) is a fundamental step in the PV pipeline, similarly to  the well-studied Named Entity Recognition (NER) task for general natural language processing (NLP). DNR aims to find drug mentions in unstructured biomedical texts and classify them into predefined categories in order to link drug names with their effects and explore drug-drug interactions (DDIs). Conventional  approaches to DNR sub-divide as rule-based, dictionary-based and machine learning-based. Intrinsically, rule-based systems are hard to scale, time-consuming to assemble and ineffective in the presence of informal sentences and abbreviated phrases. Dictionary-based systems identify drug names by matching text chunks against drug dictionaries. These systems typically achieve high precision, but suffer from low recall (i.e., they miss a significant number of mentions) due to spelling errors or drug name variants not present in the dictionaries~\cite{liu2015drug}. Conversely, machine-learning approaches have the potential to overcome all these limitations since their foundations are intrinsically robust to variants. The current state-of-the-art machine learning approaches follow a two-step process of feature engineering and classification~\cite{segura2015exploring,abacha2015text,huber2013wbi}. Feature engineering refers to the task of representing text by dedicated numeric vectors using domain knowledge. Similarly to the design of rule-based systems, this task requires much expert knowledge, is typically challenging and time-consuming, and has a major impact on the final accuracy. For this reason, this paper explores the performance of contemporary recurrent neural networks (RNNs) at providing end-to-end DNR straight from text, without any manual feature engineering stage. The tested RNNs include the popular Elman and Jordan networks and the bidirectional long short-term memory (LSTM) with decoding provided by a conditional random field (CRF)~\cite{elman1990finding,jordan1986serial,lample2016neural,collobert2011natural}. The experimental results over the SemEval-2013 Task 9.1 benchmarks show an interesting accuracy from the LSTM-CRF that exceeds that of various manually-engineered systems and approximates the best result in the literature. 

\section{Related Work}
Most of the research on drug name recognition to date has focussed on domain-dependent aspects and specialized text features. The benefit of leveraging such tailored features was made evident by the results from the SemEval-2013 Task 9.1 (Recognition and classification of pharmacological substances, known as DNR task) challenge. The system that ranked first, WBI-NER~\cite{huber2013wbi}, adopted very specialized features derived from an improved version of the ChemSpot tool~\cite{rocktaschel2012chemspot}, a collection of drug dictionaries and ontologies. Similarly, many other recent approaches~\cite{abacha2015text,liu2015feature,segura2015exploring} have been based on various combinations of general and domain-specific features. In the broader field of machine learning, the recent years have witnessed a rapid proliferation of deep neural networks, with unprecedented results in tasks as diverse as visual, speech and named-entity recognition \cite{hinton2012deep,krizhevsky2012imagenet,lample2016neural}. One of the main advantages of neural networks is that they can learn the feature representations automatically from the data, thus avoiding the laborious feature engineering stage~\cite{mesnil2015using,lample2016neural}. Given these promising results, the main goal of this paper is to provide the first performance investigation of popular RNNs such as the Elman and Jordan networks and the bidirectional LSTM-CRF over DNR tasks.

\begin{table*}[ht]
	\small
	\centering
	\scalebox{0.9}{
		\begin{tabular}{|c|c|c|c|c|c|c|c|c|c|}
			\hline \bf Sentence & \textit{Cimetidine}& \textit{reduces} & \textit{clearance}& \textit{of} & \textit{ALFENTA} &\textit{and} &\textit{volatile} &\textit{inhalation} &\textit{anesthetics}  \\ 
			\hline \textbf{Entity class}& \textit{B-drug}& \textit{O}& \textit{O} & \textit{O}& \textit{B-brand} & \textit{O} & \textit{B-group } & \textit{I-group} & \textit{I-group}  \\ 			
			\hline
		\end{tabular}}
		\caption{Example sentence in a DNR task with entity classes represented in IOB format.}
		\label{table1} 
\end{table*}
	
\begin{table*}[ht]	
	\centering
	\scalebox{0.9}{
	\begin{tabular}{|c|c|c|c|c|}
		\hline
		\multirow{3}{*}{} & 
		\multicolumn{2}{c|}{\bf {\small DDI-DrugBank}} & 
		\multicolumn{2}{c|}{\bf {\small DDI-MedLine}} \\  
		\cline{2-5}
		\cline{2-5}
		& Training+Test for DDI task & Test for DNR & Training+Test for DDI task & Test for DNR\\		
		\hline
		documents &$730$ &$54$ & $175$ & $58$ \\
		sentences &$6577$&$145$ &$1627$  & $520$\\ 
		\hline		
		drug\_n  & $124$   & $6$ &  $520$ & $115$ \\		
	    group&  $3832$ & $65$&  $234$ & $90$\\		
		brand& $1770$ & $53$ &  $36$ & $6$\\
		drug&  $9715$ & $180$& $1574$ & $171$\\  \hline
	\end{tabular}}	
	\caption{Statistics of training and test datasets used for SemEval-2013 Task 9.1.}
	\label{table2} 
\end{table*}
	
\section{The Proposed Approach}
DNR can be formulated as a joint segmentation and classification task over a predefined set of classes. As an example, consider the input sentence provided in Table~\ref{table1}. The notation follows the widely adopted in/out/begin (IOB) entity representation with, in this instance, \textit{Cimetidine} as the drug, \textit{ALFENTA} as the brand, and words \textit{volatile inhalation anesthetics} together as the group. In this paper, we approach the DNR task by recurrent neural networks and we therefore provide a brief description hereafter. In an RNN, each word in the input sentence is first mapped to a random real-valued  vector of arbitrary dimension, $d$. Then, a measurement for the word, noted as $x(t)$, is formed by concatenating the word's own vector with a window of preceding and following vectors (the "context''). An example of input vector with a context window of size $s = 3$ is:

\vspace{-0.7 cm}

\begin{equation}
\begin{split}
w_{3}(t) = [Cimetidine, \textbf{reduces}, effect], \\
`reduces' \rightarrow x_{reduces} \in \mathbb{R}^{d}, \\
`Cimetidine' \rightarrow x_{Cimetidine} \in \mathbb{R}^{d}, \\
`effect' \rightarrow x_{effect} \in \mathbb{R}^{d}, \\
x(t) = [x_{Cimetidine}, x_{\textbf{reduces}}, x_{effect}] \in \mathbb{R}^{3d}
\end{split}
\end{equation}

\noindent where $w_{3}(t)$ is the context window centered around the $t$-th word, $'reduces'$, and $x_{word}$ represents the numerical vector for $word$. 

For the Elman network, both $x(t)$ and the output from the hidden layer at time $t-1$, $h(t -1)$, are input into the hidden layer for frame $t$. The recurrent connection from the past time frame enables a short-term memory, while hidden-to-hidden neuron connections make the network Turing-complete. This architecture, common in RNNs, is suitable for prediction of sequences. Formally, the hidden layer is described as:   

\vspace{-0.45 cm}
\begin{equation}
 h(t) = f(U \bullet x(t) + V \bullet h(t-1) )
\end{equation}

\noindent where $U$ and $V$ are randomly-initialized weight matrices between the input and the hidden layer, and between the past and current hidden layers, respectively. Function $f(\cdot)$ is the sigmoid function:
   
\begin{equation}
f(x)=\frac{1}{1+e^{-x}}
\end{equation} 

\noindent that adds non-linearity to the layer. Eventually, $h(t)$ is input in the output layer:

\vspace{-0.8 cm}

\begin{equation}
\label{eq4}
y(t) = g(W \bullet h(t)), \hspace{0.03in} \text{with} \hspace{0.03in}  g(z_{m}) = \frac{e^{z_{m}}} {\Sigma _{k=1}^Ke^{z_{k}} }
\end{equation}

\noindent and convolved with the output weight matrix, $W$. The output is normalized by a multi-class logistic function, $g(\cdot)$, to become a proper probability over the class set. The output dimensionality is therefore determined by the number of entity classes (i.e., $4$ for the DNR task).The Jordan network is very similar to the Elman network, except that the feedback is sourced from the output layer rather than the previous hidden layer:

\vspace{-0.5 cm}
\begin{equation}
h(t) = f( U \bullet x(t) + V \bullet y(t-1) ).
\end{equation}

Although the Elman and Jordan networks can learn long-term dependencies, their exponential decay biases them toward their most recent inputs~\cite{bengio1994learning}. The LSTM was designed to overcome this limitation by incorporating a gated memory-cell to capture long-range dependencies within the data~\cite{hochreiter1997long}. In the bidirectional LSTM, for any given sentence, the network computes both a left, $\overrightarrow{h}(t)$, and a right, $\overleftarrow{ h}(t)$, representations of the sentence context at every input, $x(t)$. The final representation is created by concatenating them as $h(t) = [\overrightarrow{h}(t)$;$\overleftarrow{ h}(t)]$. All these networks utilize the $h(t)$ layer as an implicit feature for entity class prediction: although this model has proved effective in many cases, it is not able to provide joint decoding of the outputs in a Viterbi-style manner (e.g., an I-group cannot follow a B-brand; etc). Thus, another modification to the bidirectional LSTM is the addition of a conditional random field (CRF)~\cite{lafferty2001conditional} as the output layer to provide optimal sequential decoding. The resulting network is commonly referred to as the bidirectional LSTM-CRF \cite{lample2016neural}.

\begin{table*}[ht]	
	\centering
	\scalebox{0.89}{
	\begin{tabular}{|c|c|c|c|c|c|c|}
		\hline
		\multirow{3}{*}{Methods} & 
		\multicolumn{3}{c|}{\bf {\small DDI-DrugBank}} & 
		\multicolumn{3}{c|}{\bf {\small DDI-MedLine}} \\  
		\cline{2-7}
		\cline{2-7}
		& Precision & Recall & F$_1$ Score & Precision & Recall & F$_1$ Score\\
		\hline
		WBI-NER \cite{huber2013wbi} &$88.00$ &$87.00$ & $87.80$& $61.00$&$56.00$&$58.10$ \\
		Hybrid-DDI \cite{abacha2015text} &$93.00$&$70.00$ &$80.00$ & $74.00$& $25.00$&$37.00$ \\	
		Word2Vec+DINTO \cite{segura2015exploring}&$69.00$&$82.00$ & $75.00$& $65.00$& $51.00$&$57.00$ \\		
		\hline
		Elman RNN &$79.91$ &$60.91$ & $69.13$& $43.23$& $33.56$&$37.78$ \\
		Jordan RNN &$77.59$ &$60.91$ & $68.25$& $59.47$& $30.20$&$40.06$ \\
		Bidirectional LSTM-CRF &$87.07$ &$83.39$ & $85.19$& $52.93$& $52.57$&$52.75$ \\
		\hline	
	\end{tabular}}
	\caption{Performance comparison between the recurrent neural networks (bottom three lines) and state-of-the-art systems (top three lines) over the SemEval-2013 Task 9.1.}
	\label{table3}
\end{table*}

\begin{table*}[ht]	
	\centering
	\scalebox{0.89}{
	\begin{tabular}{|c|c|c|c|c|c|c|c|}
		\hline
		\multirow{8}{*}{Bidirectional LSTM-CRF} & 
		\multirow{2}{*}{Entities} & 
		\multicolumn{3}{c|}{\bf {\small DDI-DrugBank}} & 
		\multicolumn{3}{c|}{\bf {\small DDI-MedLine}} \\ 
		\cline{3-8}
		& & Precision & Recall & F$_1$ Score & Precision & Recall & F$_1$ Score \\
		\hline
		\multirow{4}{*}{}
		&group   & $76.92$ & $90.91$ & $83.33$& $59.52$& $53.76$&$56.50$ \\
		&drug     & $90.59$ & $84.62$ & $87.50$& $65.22$& $61.05$&$63.06$ \\
		&brand   & $91.30$ & $79.25$ & $84.85$& $0.0$& $0.0$&$0.0$ \\
		&drug\_n& $0.0$ & $0.0$ & $0.0$& $40.20$& $45.45$&$42.67$ \\
		\hline		
\end{tabular}}
		\caption{SemEval-2013 Task 9.1 results by entity for the bidirectional LSTM-CRF.} 
	\label{table4} 
\end{table*}

\section{Experiments}

\subsection{Datasets}
The DDIExtraction 2013 shared task challenge from SemEval-2013 Task 9.1~\cite{segura2013semeval} has provided a benchmark corpus for DNR and DDI extraction. The corpus contains manually-annotated pharmacological substances and drug-drug interactions (DDIs) for a total of $18,502$ pharmacological substances and $5,028$ DDIs. It collates two distinct datasets: DDI-DrugBank and DDI-MedLine~\cite{herrero2013ddi}. Table~\ref{table2} summarizes the basic statistics of the training and test datasets used in our experiments. For proper comparison, we follow the same settings as \cite{segura2015exploring}, using the training data of the DNR task along with the test data for the DDI task for training and validation of DNR. We split this joint dataset into a training and validation sets with approximately $70\%$ of sentences for training and the remaining for validation. 

\subsection{Evaluation Methodology}
Our models have been blindly evaluated on unseen DNR test data using the \textit{strict} evaluation metrics. With this evaluation, the predicted entities have to match the ground-truth entities exactly, both in boundary and class. To facilitate the replication of our experimental results, we have used a publicly-available library for the implementation\footnote{\tt https://github.com/raghavchalapathy/dnr} (i.e., the Theano neural network toolkit \cite{bergstra2010theano}). The experiments have been run over a range of values for the hyper-parameters, using the validation set for selection~\cite{bergstra2012random}. The hyper-parameters include the number of hidden-layer nodes, $H \in \{25, 50, 100\}$, the context window size, $s \in \{1, 3, 5\}$, and the embedding dimension, $d \in \{50, 100, 300, 500, 1000\}$. Two additional parameters, the learning and drop-out rates, were sampled from a uniform distribution in the range $[0.05, 0.1]$. The embedding and initial weight matrices were all sampled from the uniform distribution within range $[-1, 1]$. Early training stopping was set to $100$ epochs to mollify over-fitting, and the model that gave the best performance on the validation set was retained. The accuracy is reported in terms of micro-average F$_1$ score computed using the CoNLL score function~\cite{Nadeau:07}.

\subsection{Results and Analysis}
\label{ssec:results and analysis}
Table \ref{table3} shows the performance comparison between the explored RNNs and state-of-the-art DNR systems. As an overall note, the RNNs have not reached the same accuracy as the top system, WBI-NER \cite{huber2013wbi}. However, the bidirectional LSTM-CRF has achieved the second-best score on DDI-DrugBank and the third-best on DDI-MedLine. These results seem interesting on the ground that the RNNs provide DNR straight from text rather than from manually-engineered features. Given that the RNNs learn entirely from the data, the better performance over the DDI-DrugBank dataset is very likely due to its larger size. Accordingly, it is reasonable to expect higher relative performance should larger corpora become available in the future. Table \ref{table4} also breaks down the results by entity class for the bidirectional LSTM-CRF. The low score on the $brand$ class for DDI-MedLine and on the $drug\_n$ class (i.e., active substances not approved for human use) for DDI-DrugBank  are likely attributable to the very small sample size (Table \ref{table2}). This issue is also shared by the state-of-the-art DNR systems. 

\section{Conclusion}
\label{sec:blind}
This paper has investigated the effectiveness of recurrent neural architectures, namely the Elman and Jordan networks and the bidirectional LSTM-CRF, for drug name recognition. The most appealing feature of these architectures is their ability to provide end-to-end recognition straight from text, sparing effort from laborious feature construction. To the best of our knowledge, ours is the first paper to explore RNNs for entity recognition from pharmacological text. The experimental results over the SemEval-2013 Task 9.1 benchmarks look promising, with the bidirectional LSTM-CRF ranking closely to the state of the art. A potential way to  further improve its performance would be to initialize its training with unsupervised word embeddings such as Word2Vec~\cite{Mikolov:13} and GloVe~\cite{Pennington:14}. This approach has proved effective in many other domains and still dispenses with expert annotation effort; we plan this exploration for the near future. 

\newpage 
\bibliography{emnlp2016}
\end{document}